\documentclass[conference]{IEEEtran}
\usepackage{adjustbox}
\usepackage{amssymb,amsmath}
\usepackage{subcaption}
\usepackage{graphicx}
\ifCLASSINFOpdf
\else
\fi
\begin{document}
%
\title{Clustering and Unsupervised Anomaly Detection with $l_2$ Normalized Deep Auto-Encoder Representations}

\author{\IEEEauthorblockN{Caglar Aytekin, Xingyang Ni, Francesco Cricri and Emre Aksu}
\IEEEauthorblockA{Nokia Technologies, Tampere, Finland\\
Corresponding Author e-mail: caglar.aytekin@nokia.com}
}


%


\maketitle

\begin{abstract}
Clustering is essential to many tasks in pattern recognition and computer vision. With the advent of deep learning, there is an increasing interest in learning deep unsupervised representations for clustering analysis. Many works on this domain rely on variants of auto-encoders and use the encoder outputs as representations/features for clustering. In this paper, we show that an $l_2$ normalization constraint on these representations during auto-encoder training, makes the representations more separable and compact in the Euclidean space after training. This greatly improves the clustering accuracy when $k$-means clustering is employed on the representations. We also propose a clustering based unsupervised anomaly detection method using $l_2$ normalized deep auto-encoder representations. We show the effect of $l_2$ normalization on anomaly detection accuracy. We further show that the proposed anomaly detection method greatly improves accuracy compared to previously proposed deep methods such as reconstruction error based anomaly detection. 
\end{abstract}


%
\IEEEpeerreviewmaketitle

\section{Introduction}
Cluster analysis is essential to many applications in computer vision and pattern recognition. 
Given this fact and the recent advent of deep learning, there is an increasing interest in learning deep unsupervised representations for clustering analysis \cite{DEC,IDEC,GMVAE,DCEC}. 
Most of the methods that perform clustering on deep representations, make use of auto-encoder representations (output of the encoder part) and define clustering losses on them. 
The focus of previous works have been on the choice of the auto-encoder type and architecture and the clustering loss. 
In DEC \cite{DEC}, first a dense auto-encoder is trained with minimizing reconstruction error. 
Then, as a clustering optimization stage, the method iterates between computing an auxiliary target distribution from auto-encoder representations and minimizing the Kullback-Leibler divergence to it. 
In IDEC \cite{IDEC}, it is argued that the clustering loss of DEC corrupts the feature space, therefore IDEC proposes to jointly optimize the clustering loss and reconstruction loss of the auto-encoder. 
DCEC \cite{DCEC} argues the inefficiency of using dense auto-encoders for image clustering, therefore adopts a convolutional auto-encoder and shows that it improves the clustering accuracy of DEC and IDEC. 
GMVAE \cite{GMVAE} adopts a variational auto-encoder in order to learn unsupervised representations and simply applies K-means clustering on representations.

In this manuscript, we show that regardless of the auto-encoder type (dense or convolutional), constraining the auto-encoder representations to be on the unit-ball, i.e. to be $l_2$ normalized, during auto-encoder training, greatly improves the clustering accuracy. 
We show that a simple $k$-means clustering on the auto-encoder representations trained with our constraint already gives improved accuracy with a large margin compared to baselines with or without additional clustering losses.
Motivated by the high performance of our clustering method on deep representations, we propose an unsupervised anomaly detection method based on this clustering.
We show that our anomaly detection method greatly improves on other deep anomaly detection strategies such as reconstruction error based ones.
We also investigate the effect of $l_2$ normalization constraint during training on the anomaly detection accuracy and show that it leads to superior results compared to not applying the constraint.

\section{Related Work}
\subsection{Deep Unsupervised Anomaly Detection}

Unsupervised anomaly detection tries to find anomalies in the data without using any annotation \cite{ANOM}. 
Recently, deep learning methods have also been used for this task \cite{DRAE, AVAE}. These works train auto-encoders on the entire data and use reconstruction loss as an indicator of anomaly. 
DRAE \cite{DRAE} trains auto-encoders and uses reconstruction error as an anomaly indicator. 
Moreover, DRAE proposes a method to make the reconstruction error distributions of the normal and abnormal classes even more separable so that it is easier to detect anomalies. 
AVAE \cite{AVAE} trains both conventional and variational auto-encoders and use reconstruction error as an anomaly indicator. 

The general assumption of the above works is that since the anomaly data is smaller in ratio than the normal data, the auto-encoder would not learn to reconstruct it accurately. 

The above assumption seems to work in a specific definition of anomaly where the normal samples are drawn from a single class only and anomaly classes have been selected from many other classes \cite{DRAE}.
However, the assumption fails in another anomaly type where the normal samples are drawn from multiple classes and anomaly class is sampled from a specific class \cite{AVAE}.

In this paper we propose an unsupervised anomaly detection method based on clustering on deep auto-encoder representations and show that it gives a superior performance than reconstruction error based anomaly.

\begin{figure*}

\centering
  \subcaptionbox{No normalization\label{fig1:a}}{\includegraphics[width=2.3in]{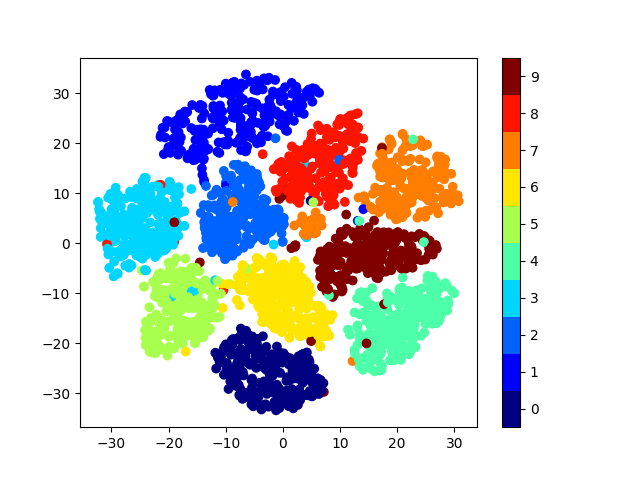}}
  \subcaptionbox{$l_2$ normalization after training\label{fig1:b}}{\includegraphics[width=2.3in]{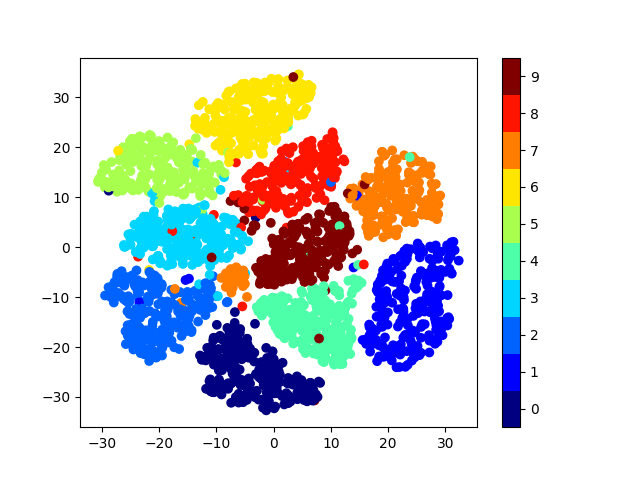}}
  \subcaptionbox{$l_2$ normalization during training\label{fig1:c}}{\includegraphics[width=2.3in]{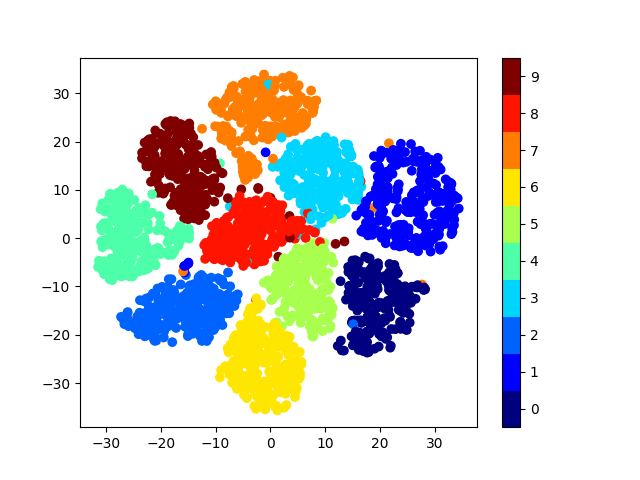}}
  
\caption{Illustration of t-SNE encoding of auto-encoder representations for MNIST dataset to two dimensions. Best viewed in color.} \label{fig:1}
\end{figure*}

\subsection{Regularization and Normalization in Neural Networks}

Due to the high number of parameters, neural networks have a risk of over-fitting to the training data. 
This sometimes reduces the generalization ability of the learned network.
In order to deal with over-fitting, mostly regularization methods are employed.
One of the most widely used regularization technique is weight norm regularization. 
Here the aim is to add an additional regularization loss to the neural network error, which gives high penalty to weights that have high norms.
Both $l_1$ and $l_2$ norm can be exploited.

Recently some normalization techniques for neural networks emerged such as \cite{BNORM, LNORM}.
Batch normalization \cite{BNORM}, aims to find a statistical mean and variance for the activations which are calculated and updated according to batch statistics.
The activations are normalized according to these statistics.
In layer normalization \cite{LNORM}, the mean and variance are computed from all of the summed inputs to the neurons on a layer on a single training sample. 
This overcomes the batch-size dependency drawback of batch-normalization.
Although these methods were mainly proposed as tricks to make the neural network training faster by conditioning each layer's input, it is argued that they may also have a regularization effect due to their varying estimations of parameters for standardization at each epoch.

In our proposed method, the unit ball constraint that we put on activations is a normalization technique.
However, unlike layer or batch normalization, the unit ball constraint is parameter-free as it simply sets the norm of each activation vector to 1.
Therefore, it is free from the parameter estimation stochasticity.
Yet, it may still act as a regularization method due to its hard constraint on some activations to be of fixed norm.
This slightly resembles the $l_2$ norm regularization.
A key difference is that in $l_2$ norm regularization, the norm is of the weights, but in our case it is applied on the activations.
Another key difference is that we fix the activation norms to 1, whereas $l_2$ norm regularization penalizes to large weight norms and does not fix the norms to any value.

\section{Proposed Method}

\subsection{Clustering on $l_2$ Normalized Deep Auto-Encoder Representations}

We represent the auto-encoder representations for the input $I$ as $E(I)$ and the reconstructed input as the $D(E(I))$. The representations are generally obtained via several dense or convolutional layers applied on the input $I$. In each layer, usually there are filtering and an activation operations, and optionally pooling operations. Let $f_i$ be the computations applied to the input at layer $i$, then the encoded representations for an $n$-layer encoder are obtained as in Eq. \ref{eqn1}.

\begin{equation}
\label{eqn1}
E(I)=f_n(f_{n-1}(...f_1(I)))
\end{equation}

The reconstruction part of the auto-encoder applies on $E(I)$ and is obtained via several dense or deconvolutional layers. In each layer, usually there are filtering and activation operations and optionally un-pooling or up-sampling operations. Let $g_i$ be the computations applied to the auto-encoder representations, then the reconstructed signal for an $m$-layer decoder is obtained as as in Eq. \ref{eqn2}.

\begin{equation}
\label{eqn2}
D(E(I))=g_m(g_{m-1}(...g_1(E(I))))
\end{equation}

The auto-encoder training is conducted in order to reduce the reconstruction error (loss) given in Eq. \ref{eqn3}.

\begin{equation}
\label{eqn3}
L=\frac{1}{|J|}\sum_{j \in J} (I_j-D(E(I_j)))^2
\end{equation}

%
%

Here, we propose an additional step conducted on the auto-encoder representations $E(I)$. 
In particular, we apply $l_2$ normalization on $E(I)$. 
This corresponds to adding a hard constraint on the representations to be on the unit ball.
The loss function with our introduced constraint can then be written as in Eq. \ref{eqn4}, where $L_c$ and $E_c$ are loss and encoded representations with our introduced constraint.

\begin{figure*}

\centering
  \subcaptionbox{Clustering Method\label{method:a}}{\includegraphics[width=3.2in]{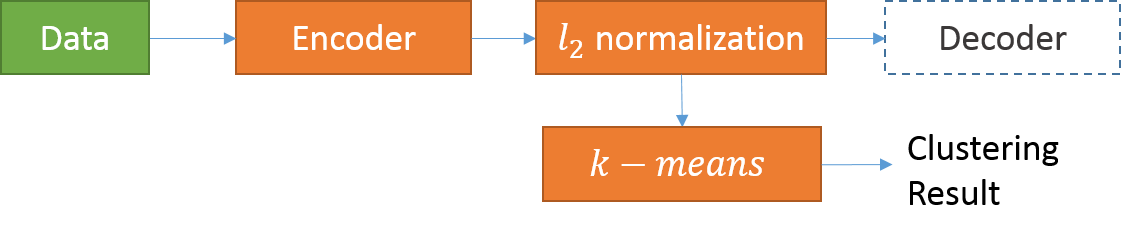}}
  \hfill
\subcaptionbox{Normality Score Method\label{method:b}}{\includegraphics[width=3.6in]{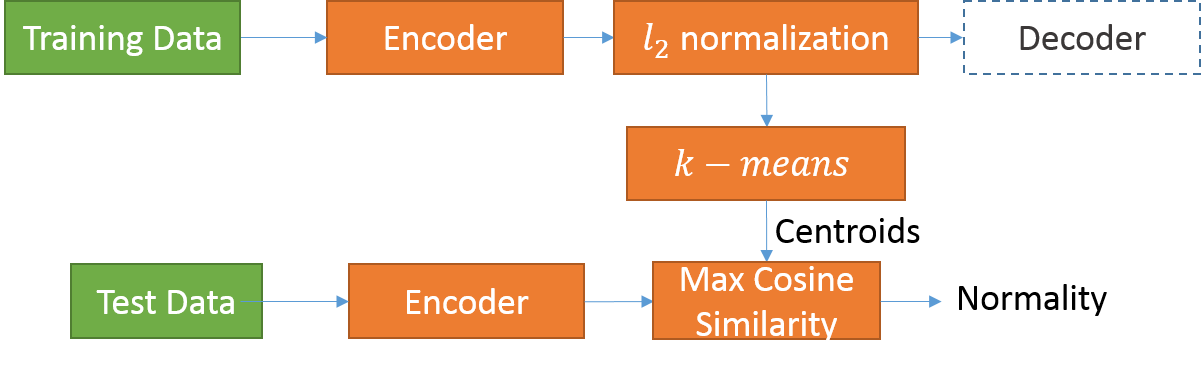}}
  
\caption{Illustration of proposed methods in inference phase.}
\end{figure*}

\begin{equation}
\label{eqn4}
\begin{aligned}
L_c=\frac{1}{|J|}\sum_{j \in J} (I_j-D(E_c(I_j)))^2 ,\\
E_c(I)=\frac{E(I)}{\lVert E(I) \rVert_2}
\end{aligned}
\end{equation}

We believe that $l_2$ normalized features (representations) are more suitable for clustering purposes, especially for the methods that use Euclidean distance such as conventional $k$-means. 
This is because the distances between the vectors would be independent of their length, and would instead depend on the angle between the vectors. 
As a positive side effect, enforcing the unit norm on representations would act as a regularization for the entire auto-encoder.

In Fig. \ref{fig:1}, we illustrate t-SNE \cite{TSNE} encoding (to 2 dimensions) of the auto-encoder representations of networks with same architecture. 
All auto-encoders were trained on MNIST \cite{MNIST} dataset training split. 
The representations corresponding to Fig. \ref{fig1:a} are from the auto-encoder that was trained by the loss in Eq. \ref{eqn3}.
The same representations with $l_2$ normalization applied after training are illustrated in Fig. \ref{fig1:b}.
Finally, the representations with $l_2$ constraint during training , i.e. training with loss Eq. \ref{eqn4}, are illustrated in Fig. \ref{fig1:c}.

It is observed from the figures that the $l_2$ normalization during training (Fig. \ref{fig1:c}) results into more separable clusters. One example is the distributions of digit 7 in MNIST dataset. Note that the numbers are indicated with color codes where the color bar is available in Fig. \ref{fig:1}. It is clearly observed that with no normalization during training (Fig. \ref{fig1:a}), digit 7 is divided into 2 parts where the small part is surrounded by 8,9,6 and 2 digits. With normalization applied after training (Fig. \ref{fig1:b}) this effect becomes even more evident. So, we clearly observe here that applying normalization after training does not help at all. But, with $l_2$ normalization constraint during training (Fig. \ref{fig1:c}), we see a clear separation of digit 7 as a single cluster from the rest of the numbers. Moreover, it can be observed that the clusters are more compact in Fig. \ref{fig1:c} compared to others.

After training the auto-encoder with the loss function in Eq. \ref{eqn4}, the clustering is simply performed by $k$-means algorithm. No more clustering loss is applied. Our clustering method is illustrated in Fig. \ref{method:a}.

\subsection{Unsupervised Anomaly Detection using $l_2$ Normalized Deep Auto-Encoder Representations} 
\label{anomalymethod}
Here, we propose a clustering based unsupervised anomaly detection.
We train an auto-encoder on the entire dataset including normal and abnormal samples and no annotation or supervision is used.
The auto-encoder is simply trained with the loss in Eq. \ref{eqn4}.
After training, the $l_2$ normalized auto-encoder representations are clustered with $k$-means algorithm.
We assume the anomaly cases to be considerably smaller in number than any normal clusters.
Note that this assumption does not put any constraint on the dataset, but it simply follows the general definition of anomaly.
Therefore, the centroids obtained by the $k$-means method can be considered to be representations of normal clusters by some errors that are caused by anomaly samples in the dataset.
To each sample $i$, we assign the normality score $v_i$ in Eq. \ref{eqn5}.

\begin{equation}
\label{eqn5}
v_i=\max_{j} (E_c(I_i) \cdot \frac{C_j}{\lVert C_j \rVert_2})
\end{equation}
In Eq. \ref{eqn5}, $C_j$ is a cluster centroid and $\cdot$ is the dot product operator.
Notice that we $l_2$ normalize the cluster centroids.
Since representations $E_c(I_i)$ are already $l_2$ normalized, $v_i \in [0,1]$ holds.
The measure in Eq. \ref{eqn5} is intuitive considering that we expect high similarities of normal samples to normal classes.
Our normality scoring method is illustrated in Fig. \ref{method:b}.

The normality score can be used for anomaly detection in a straightforward manner.
Simply, the abnormal samples can be detected as the ones having $v_i<\tau$ , where $\tau \in [0,1]$ is a threshold.

\section{Experimental Results}
\subsection{Clustering}

\subsubsection*{Evaluation Metrics}
We use the widely used evaluation for unsupervised clustering accuracy \cite{DEC} given in Eq. \ref{acc}.
\begin{equation}
\label{acc}
acc=\max_m \frac{\sum_{i=1}^n \pmb{1} \{l_i=m(c_i) \}}{n}
\end{equation}
In Eq. \ref{acc}, $l_i$ is the ground truth labeling of sample $i$, $c_i$ is the cluster assignment according to the one-to-one mapping defined by $m$, where $m$ ranges over all possible one-to-one mappings between clusters generated by the algorithm and the ground truth labels.
The maximization in Eq. \ref{acc} can be performed by Hungarian algorithm \cite{HUN}.

We compare the clustering accuracy of auto-encoder representations with and without $l_2$ normalization constraint.
We make this comparison in dense and convolutional auto-encoders.
For dense auto-encoder, we use MNIST \cite{MNIST} and for convolutional auto-encoders, we use MNIST \cite{MNIST} and USPS \cite{USPS} datasets. 
This is due to the availability of results in the works that use dense and convolutional auto-encoders.

For dense auto-encoder, we use the network structure which is used both in DEC \cite{DEC} and IDEC \cite{IDEC}. 
In encoding part there are 4 layers with $500-500-2000-10$ hidden neurons and in decoding part there are $2000-500-500-d$ neurons, where $d$ is the dimension of the input. 
We re-implement the auto-encoder training and with leaky relu \cite{LRELU} activations after each hidden layer except for the last one and trained the auto-encoder end to end for 100 epochs. 
We select the best model with lowest reconstruction error.
As it can be observed from Table \ref{AE}, we obtain a very similar clustering accuracy when we apply $k$-means on auto-encoder representations compared to the original paper of DEC \cite{DEC}. 
Note here that results indicated with * corresponds to our own implementation.
Other results for baselines are borrowed from original papers.
Table \ref{AE} shows that when we train the auto-encoder with our $l_2$ normalization constraint on the representations, we achieve a much better clustering accuracy when we apply $k$-means on the representations. 
We denote our method as AE-$l_2$ which stands for auto-encoder with $l_2$ normalization.
Moreover, our clustering accuracy is even better than the methods that define a separately designed clustering loss on the representations (DEC and IDEC).

Next, we make experiments for the convolutional auto-encoder. 
For this, we make use of the model structure introduced in DCEC \cite{DCEC}. 
This model consists of  5x5, 5x5 and 3x3 convolutional filters in the encoding layers respectively. 
There are 32,64 and 128 filters in encoding layers respectively. 
Convolutions are applied with 2x2 strides and with relu \cite{RELU} activations. 
After the convolutional layers, the activations are flattened and there is a dense layer of dimension 10. 
This is followed by another dense layer and reshaping. 
Decoder part consists of 64,32 and 1 deconvolutional filters  of size 3x3, 5x5 and 5x5 respectively. 
Relu activations were applied after each convolution, except for the last one.
The network was trained for 200 epochs as in the original paper of DCEC \cite{DCEC}. 
In Table \ref{CAE}, we show clustering accuracy of $k$-means applied on convolutional autoencoder representations. 
We were able to obtain similar results as in the original paper (DCEC). 
Note here that results indicated with * corresponds to our own implementation.
Other results for baselines are borrowed from original papers.
It can be observed from Table \ref{CAE} that when we train the convolutional autoencoder with our $l_2$ normalization constraint on representations, we achieve a much better performance. 
We denote our method as CAE-$l_2$ which stands for convolutional auto-encoder with $l_2$ normalization.
Our performance is superior to DCEC which introduces additional clustering loss.

\begin{table}[!t]

\caption{Clustering on Dense Auto-Encoder Representations}
\label{AE}
\centering
\begin{adjustbox}{width=0.47\textwidth}

\begin{tabular}{|c|c|c|c|c|c|}
\hline
&AE* & AE & DEC & IDEC & AE-$l_2$\\ 
&$k$-means & $k$-means &  &  & $k$-means\\ 
\hline
MNIST&81.43 & 81.82 & 86.55 & 88.06 & \textbf{90.20}\\

\hline
\end{tabular}
\end{adjustbox}
\end{table}

\begin{table}[!t]

\caption{Clustering on Convolutional Auto-Encoder Representations}
\label{CAE}
\centering
\begin{adjustbox}{width=0.4\textwidth}

\begin{tabular}{|c|c|c|c|c|}
\hline
&CAE* & CAE & DCEC & CAE-$l_2$\\ 
&$k$-means & $k$-means &  & $k$-means\\ 

\hline
MNIST&84.83 & 84.90 & 88.97 & \textbf{95.11}\\
\hline
USPS&73.521 & 74.15 & 79.00 & \textbf{91.35}\\
\hline
\end{tabular}
\end{adjustbox}

\end{table}

\
\subsubsection*{$l_2$ versus Batch and Layer Normalization}

Due to $l_2$ normalization step in our clustering method, we compare it with applying other normalization techniques training. 
In particular we train two separate networks by using batch \cite{BNORM} and layer \cite{LNORM} normalization, instead of $l_2$ normalization.
All other setup for the experiments are the same.
Batch size of 256 is used for all methods in order to have a large enough batch for batch normalization.
Our method performs superior to both baselines by a large margin, as the accuracies in Table \ref{regul} indicate.
More importantly it is noticed that neither batch nor layer normalization provides a noticeable accuracy increase over the baseline (CAE+$k$-means).
Moreover in MNIST dataset, layer and batch normalization results into a significant accuracy decrease.
This is an important indicator showing that the performance upgrade of our method is not a result of a input conditioning, but it is a result of the specific normalization type that is more fit for clustering in Euclidean space.

\begin{table}[!t]

\caption{Comparison of Normalization Methods}
\label{regul}
\centering
\begin{adjustbox}{width=0.4\textwidth}

\begin{tabular}{|c|c|c|c|}
\hline
 & batch-norm & layer-norm & $l_2$-norm\\ 
\hline
MNIST & 70.67  & 70.83 & \textbf{95.11}\\
\hline
USPS & 74.95 & 75.263 & \textbf{91.35}\\
\hline
\end{tabular}
\end{adjustbox}
\end{table}

\subsection{Anomaly Detection}

\subsubsection*{Evaluation Metrics}
An anomaly detection method often generates an anomaly score, not a hard classification result. 
Therefore, a common evaluation strategy in anomaly detection is to threshold this anomaly score and form a receiver operating curve where each point is the true positive and false positive rate of the anomaly detection result corresponding to a threshold. 
Then, the area under the curve (AUC) of RoC curve is used as an evaluation of the anomaly detection method \cite{ANOM}. 

Here, we evaluate our method introduced in Section \ref{anomalymethod}.
The evaluation setup and implementation of our method are as follows.
In MNIST training dataset, we select a digit class as anomaly class and keep a random 10$\%$ of that class in the dataset while the remaining 90$\%$ is ignored. 
We leave the rest of the classes as is. 
Then, we use the convolutional autoencoder structure in DCEC \cite{DCEC} and train it with our $l_2$ normalization constraint on representations. 
Finally, we apply $k$-means clustering on the representations and keep the centroids. 
In our experiments we use k=9 for $k$-means, since we assume that we know the number of normal classes in the data. 
For MNIST test dataset, we calculate the auto-encoder representations. 
As a normality measure for each sample, we calculate the corresponding representation's maximum similarity to pre-calculated cluster centroids as in \ref{eqn5}.
It should be noted that we repeat the above procedure by choosing a different class to be anomaly class, for all possible classes. 

We also evaluate two baselines. 
In the first baseline, we exactly repeat the above procedure, but without $l_2$ normalization constraint on representations. 
In the second baseline, again we train the auto-encoder with $l_2$ normalization constraint on representations.
Then, on the test set, we calculate the reconstruction error per sample and define that as anomaly score.
Using reconstruction error based anomaly detection follows the works in AVAE \cite{AVAE} and DRAE \cite{DRAE}.
The setups in AVAE and DRAE are different than ours. 
In AVAE, the training is only conducted on normal data, so the method is not entirely unsupervised in that sense.
In DRAE, the anomaly definition is different: only a single class is kept as normal and samples from other classes are treated as anomaly.
That setup presents a much easier case and therefore reconstruction error based anomaly detection produces acceptable results.
Next, we show that in our setup this is not the case.

For each method we plot a RoC curve via thresholding the normality (or anomaly) score with multiple thresholds. 
Then, we evaluate the area under the RoC curve (AUC) for measuring the anomaly detection performance. 
The training and test datasets for all methods are the same. 
Due to the random selection of 10$\%$ of the anomaly class to be kept, performance can change according to the partition that is randomly selected. Therefore, we run the method 10 times for different random partitions and report the mean AUC.

It can be observed from Table \ref{anom} that our clustering based anomaly detection method drastically outperforms the reconstruction error based anomaly detection for CAE neural network structure. 

It is worth noting here an interesting observation from Table \ref{anom}: for digits 1, 7 and 9, reconstruction error based anomaly detection gets a very inferior performance. 
This is most evident in digit 1. 
The reason for this is that the digit 1 is very easy to reconstruct (only 1 stroke) and even though an auto-encoder is trained on much less examples of this digit, it can reconstruct it quite well. 
This shows a clear drawback of the reconstruction error based anomaly detection. 
However, in our clustering based method, we achieve a very high accuracy in all the digits.

The effect of our proposed $l_2$ normalization constraint on representations during training can also be observed from Table \ref{anom}.
In $9/10$ cases, i.e. digits selected as anomaly, anomaly detection with the network trained with $l_2$ normalization constraint on representations performs much better than the one without. 
Only in digit $9$, we observe an inferior accuracy of our method.
Compared to other digits, we observe less performance for digits 4 and 9. 
We argue that this might be happening due to very similar appearance of these digits in some handwritings.
Therefore, the method may confuse these numbers with each other during clustering.

\begin{table}[!t]

\caption{Anomaly Detection with Auto-Encoder Representations}
\label{anom}
\centering
\centering

\begin{adjustbox}{width=0.35\textwidth}

\begin{tabular}{|c|c|c|c|c|}
\hline
Anom.  & CAE. & CAE & CAE-$l_2$\\ 
Digit  & (recons) & (cluster) & (cluster)\\ 
\hline
0 &  0.7025 & 0.7998 & \textbf{0.9615} \\
\hline
1 &  0.0782 & 0.8871 & \textbf{0.9673}\\
\hline
2 &  0.879 & 0.7512 & \textbf{0.9790}\\
\hline
3 & 0.8324 & 0.8449 & \textbf{0.9382}\\
\hline
4 & 0.7149 & 0.4988 & \textbf{0.7825}\\
\hline
5 &  0.8359 & 0.7635 & \textbf{0.9136}\\
\hline
6 & 0.6925 & 0.7896 & \textbf{0.9497}\\
\hline
7 &  0.5767 & 0.7421 & \textbf{0.9100}\\
\hline
8 &  0.8912 & 0.9200 & \textbf{0.9237}\\
\hline
9 &  0.514 & \textbf{0.8944} & 0.7495\\
\hline
\end{tabular}
\end{adjustbox}
\end{table}

In Table \ref{anom2}, we compare our method to another method \cite{AVAE} that performs reconstruction error based anomaly detection, but using dense auto-encoders. 
There is also a variational auto-encoder based version of the method. 
It should be noted that this method trains auto-encoders only on normal data. 
This presents a much easier task compared to our case where we also include anomalous samples during training. 
Thus our case is entirely unsupervised.
Still, $9/10$ cases, our method outperforms both variants of the method with a large margin.
Only in digit 4, we observe an inferior performance of our method compared to VAE method. 

\begin{table}[!t]

\caption{Anomaly Detection with Auto-Encoder Representations}
\label{anom2}
\centering
\begin{adjustbox}{width=0.35\textwidth}

\begin{tabular}{|c|c|c|c|c|}
\hline
Anom. & AE. & VAE. & CAE-$l_2$\\ 
(Digit) & (recons.) & (recons.) & cluster\\ 
\hline
0 & 0.825 & 0.917& \textbf{0.9615} \\
\hline
1 & 0.135 & 0.136& \textbf{0.9673}\\
\hline
2 & 0.874 & 0.921  & \textbf{0.9790}\\
\hline
3 & 0.761 & 0.781  & \textbf{0.9382}\\
\hline
4 & 0.727 & \textbf{0.808} &  0.7825\\
\hline
5 & 0.792 & 0.862 &  \textbf{0.9136}\\
\hline
6 & 0.812 & 0.848 & \textbf{0.9497}\\
\hline
7 & 0.508 & 0.596 &\textbf{0.9100}\\
\hline
8 & 0.869 & 0.895  & \textbf{0.9237}\\
\hline
9 & 0.548 & 0.545  & \textbf{0.7495}\\
\hline
\end{tabular}
\end{adjustbox}

\end{table}

\section{Conclusion}
In this paper, we have applied a $l_2$ normalization constraint to deep autoencoder representations during autoencoder training and observed that the representations obtained in this way clusters well in Euclidean space.
Therefore, applying a simple $k$-means clustering on these representations gives high clustering accuracies and works better than methods defining additional clustering losses.
We have also shown that the high performance is not due to any conditioning applied on the representations but it is due to selection of a particular normalization that leads to more separable clusters in Euclidean space.
We have proposed an unsupervised anomaly detection method on $l_2$ normalized deep auto-encoder representations.
We have shown that the proposed $l_2$ normalization constraint drastically increases the anomaly detection method's performance. 
Finally, we have shown that the commonly adopted deep anomaly detection method based on the reconstruction error performs weak in a definition of anomaly, whereas our method performs superior.






%

\end{document}